\documentclass[10pt,twocolumn,letterpaper]{article}

\usepackage[pagenumbers]{cvpr} % To force page numbers, e.g. for an arXiv version

\definecolor{cvprblue}{rgb}{0.21,0.49,0.74}
\usepackage[pagebackref,breaklinks,colorlinks,allcolors=cvprblue]{hyperref}

\def\paperID{15} % *** Enter the Paper ID here
\def\confName{CVPR}
\def\confYear{2025}

\title{Efficient Egocentric Action Recognition with Multimodal Data}

\author{Marco Calzavara\\
ETH Zurich
\and
Ard Kastrati\\
ETH Zurich
\and
Matteo Macchini\\
Magic Leap
\and 
Dushan Vasilevski\\
Magic Leap
\and
Roger Wattenhofer\\
ETH Zurich
}

\begin{document}
\maketitle
\begin{abstract}
The increasing availability of wearable XR devices opens new perspectives for Egocentric Action Recognition (EAR) systems, which can provide deeper human understanding and situation awareness. However, deploying real-time algorithms on these devices can be challenging due to the inherent trade-offs between portability, battery life, and computational resources. In this work, we systematically analyze the impact of sampling frequency across different input modalities---RGB video and 3D hand pose---on egocentric action recognition performance and CPU usage. By exploring a range of configurations, we provide a comprehensive characterization of the trade-offs between accuracy and computational efficiency. Our findings reveal that reducing the sampling rate of RGB frames, when complemented with higher-frequency 3D hand pose input, can preserve high accuracy while significantly lowering CPU demands. Notably, we observe up to a 3× reduction in CPU usage with minimal to no loss in recognition performance. This highlights the potential of multimodal input strategies as a viable approach to achieving efficient, real-time EAR on XR devices.
\end{abstract}    
\section{Introduction}

Since the advent of head-mounted devices, researchers and industry have shown growing interest in leveraging video captured by wearable cameras to extract valuable information. Among these devices, AR glasses such as Magic Leap 2\footnote{https://www.magicleap.com/magic-leap-2} stand out by integrating digital content into the physical world. The data they capture can unlock numerous applications, from task assistance to personalized recommendations and contextual awareness.

Egocentric Action Recognition (EAR) is a research field focused on identifying human actions from first-person data. It presents unique challenges, including variable viewpoints, frequent occlusions, and the computational constraints of wearable devices. On AR glasses, efficiency is especially critical, not only to preserve battery life but also because these devices typically lack dedicated hardware acceleration and must rely on limited CPU resources. This project tackles these challenges by targeting efficient and accurate real-time action recognition for AR glasses. 

EAR methods are typically categorized as either uni-modal or multi-modal. Several studies rely solely on RGB data for EAR. For example, \cite{pirsiavash2012detecting} adopts a Bag-of-Objects approach, while \cite{kapidis2020object} uses frame-level features derived from object detections. Other methods extract features directly from RGB frames. Among these, \cite{Sudhakaran_2017_ICCV} combines 2D and 3D CNNs with a ConvLSTM \cite{shi2015convolutional}, also testing frame differences instead of raw frames. Although RGB-based systems perform well, some studies explore whether hand pose alone is sufficient. For instance, \cite{das2021symmetric} uses a Spatio-Temporal Graph Convolutional Network on hand skeletons, which is inspired by \cite{yan2018spatial}. These results highlight a strong link between hand motion and egocentric actions.

The main drawback of uni-modal systems is their inability to capture diverse cues, such as object identity or hand-object interactions. Multi-modal models address this limitation by combining multiple inputs. A popular example is the two-stream architecture from \cite{simonyan2014two}, which processes RGB and optical flow separately. In EAR, gaze-based attention is used in \cite{Li_2018_ECCV}, while \cite{sudhakaran2018attention} shows improved performance by adding an optical flow branch to \cite{Sudhakaran_2017_ICCV}. Beyond RGB and flow, other modalities have also been explored; for instance, \cite{shamil2025utility} uses RGB with 3D hand poses, avoiding the high computational cost of optical flow. Three-stream models extend this further: \cite{furnari2019would} combines RGB, optical flow, and object features, showing better results than any single stream. However, as noted in \cite{HoloAssist2023}, more modalities do not always lead to better performance.

Building upon previous work, our approach investigates a targeted two-stream architecture that utilizes RGB frames and 3D hand pose keypoints to predict the current action, aiming to optimize both accuracy and resource efficiency. This paper makes the following contributions:
\begin{itemize}
\item We introduce a multi-modal EAR system that combines an RGB stream comprising a Vision Transformer (ViT) feature extractor with a hand pose stream.
\item We present a systematic study of the interplay between sampling frequency, recognition accuracy, and computational cost across modalities. Our analysis reveals how adjusting the relative sampling rates of RGB and hand pose inputs can serve as a design lever to optimize EAR systems for resource-constrained devices.
\item We demonstrate that downsampling RGB input --- when complemented with higher-frequency hand pose signals --- preserves accuracy while significantly improving efficiency. Specifically, we show that maintaining hand pose input at 30 Hz while reducing RGB input to 10 Hz achieves competitive accuracy with up to a 3× reduction in CPU usage compared to full-frame-rate RGB baselines. These results highlight the practical value of modality-aware sampling strategies for efficient on-device deployment of EAR systems.
\end{itemize}
\section{Methods}

\subsection{Model}

\begin{figure}[thpb]
    \centering
    \begin{subfigure}{.95\linewidth}
        \centering
        \includegraphics[width=2.75in]{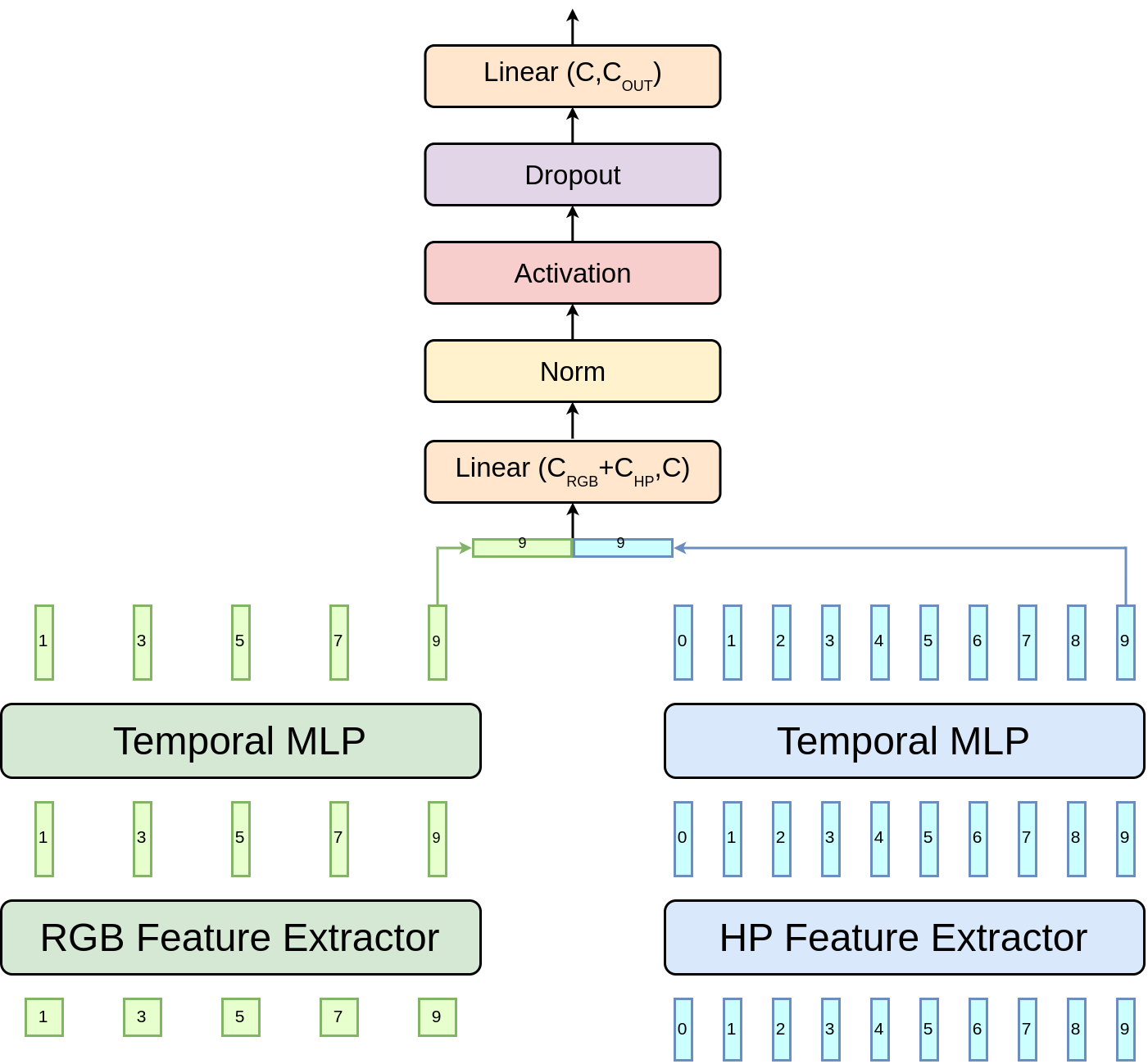}
        \caption{Architecture of the Multimodal Temporal MLP. The model consists of two parallel processing streams: one for RGB features, utilizing a LeViT feature extractor and a Temporal MLP, and another for hand pose features, employing an MLP feature extractor paired with a Temporal MLP. The outputs from both streams at the final time step are concatenated and processed through additional layers for classification.}
        \label{fig:mmtmlp}
    \end{subfigure}
    \begin{subfigure}{.95\linewidth}
        \centering
        \includegraphics[width=2.75in]{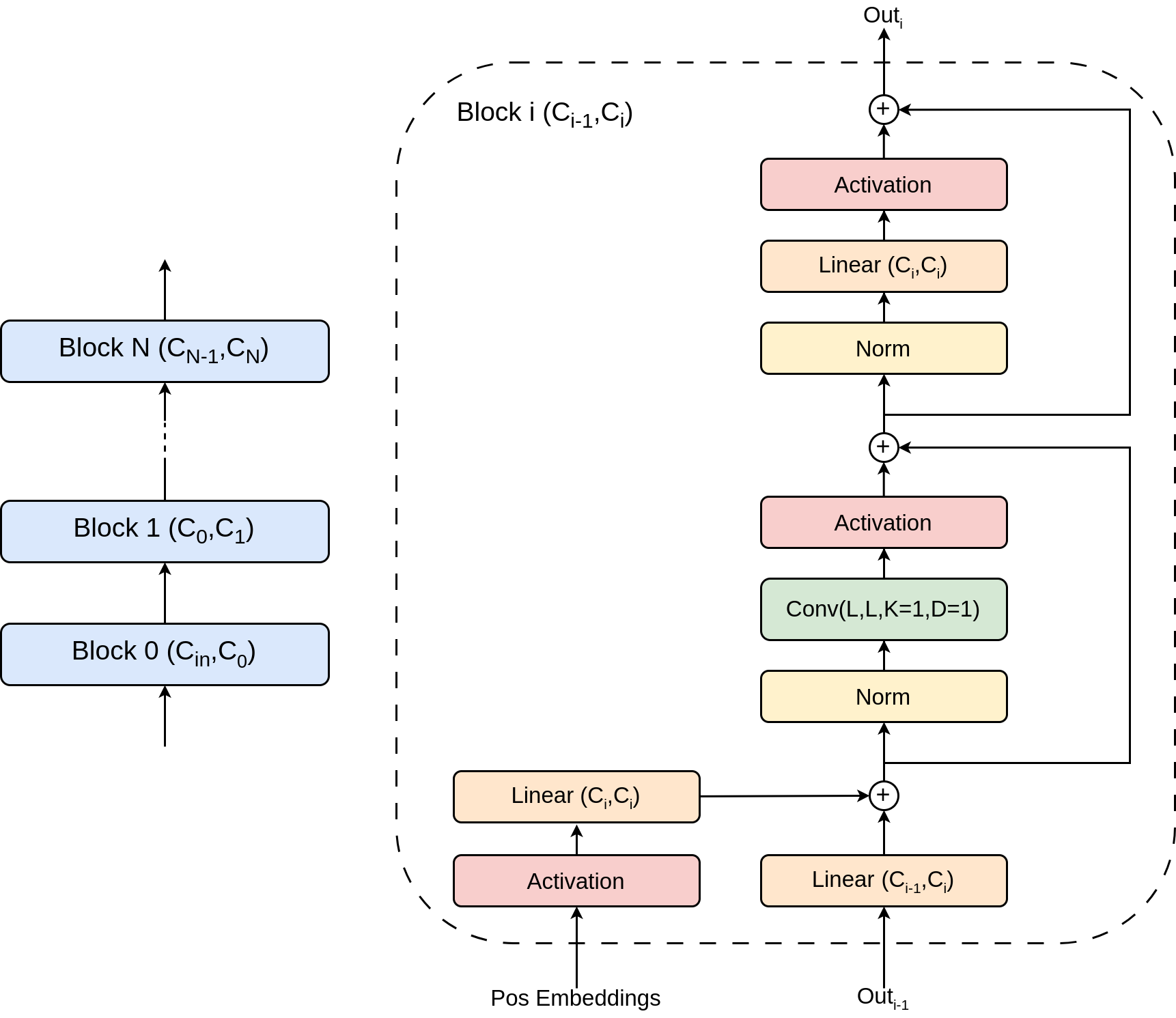}
        \caption{Architecture of the Temporal MLP. This module captures long-range dependencies in temporal sequences without relying on convolutional layers with large kernel sizes. Inspired by \cite{du2023avatars}, it employs MLP-based operations to model temporal relationships efficiently, making it well-suited for sequence-based tasks such as action recognition.}
        \label{fig:tmlp}
    \end{subfigure}
    \caption{Architectures of the Multimodal Temporal MLP and Temporal MLP.}
\end{figure}

Operating over sequences is crucial for accurate action recognition, as context from multiple consecutive time steps helps detect motion. In this work, we employ a multi-stream architecture consisting of modality-specific feature extractors and sequence models. The model, depicted in Figure \ref{fig:mmtmlp}, processes the two modalities through dedicated modules: a LeViT \cite{graham2021levit} feature extractor and Temporal MLP for RGB features and a hand pose feature extractor paired with a Temporal MLP for hand pose features. The outputs from both streams, corresponding to the final time steps, are concatenated and processed through a series of layers for classification. 

A key component of our approach is the Temporal MLP, which is responsible for modeling temporal dependencies in the feature representations. The architecture of the Temporal MLP, shown in Figure \ref{fig:tmlp}, is inspired by \cite{du2023avatars} and is designed to efficiently capture long-range dependencies without relying on convolutional layers with large kernel sizes.

\subsection{Experimental Conditions}

Model training was performed using either two or four GPUs, depending on availability. The GPUs used for training included the Titan RTX (24 GB memory), GeForce RTX 3090 (24 GB memory), and Tesla V100 (32 GB memory). For CPU usage, a single thread of an AMD EPYC 7742 CPU (base clock: 2.25 GHz) was utilized.

\subsection{Dataset}

We conducted our experiments on the training and validation sets of the H2O dataset \cite{kwon2021h2o}. While the H2O dataset is designed for predicting the action performed over a video segment, it can also be used for frame-level prediction by assigning the segment's action label to each frame within the segment.

\subsection{Preprocessing and Augmentation}

RGB frames are cropped and normalized prior to feature extraction. For 3D hand keypoints, we apply a three-step normalization to ensure spatial consistency: (1) translating the wrists to the origin, (2) standardizing edge lengths between keypoints, and (3) rotating the hands to align specific vectors with canonical axes. Beyond preprocessing, we apply several data augmentation techniques. When both RGB frames and 3D hand pose keypoints are used, only shared augmentations are applied to preserve cross-modal consistency.
\section{Results}

\begin{figure}[thpb]
    \centering
    \includegraphics[width=\linewidth]{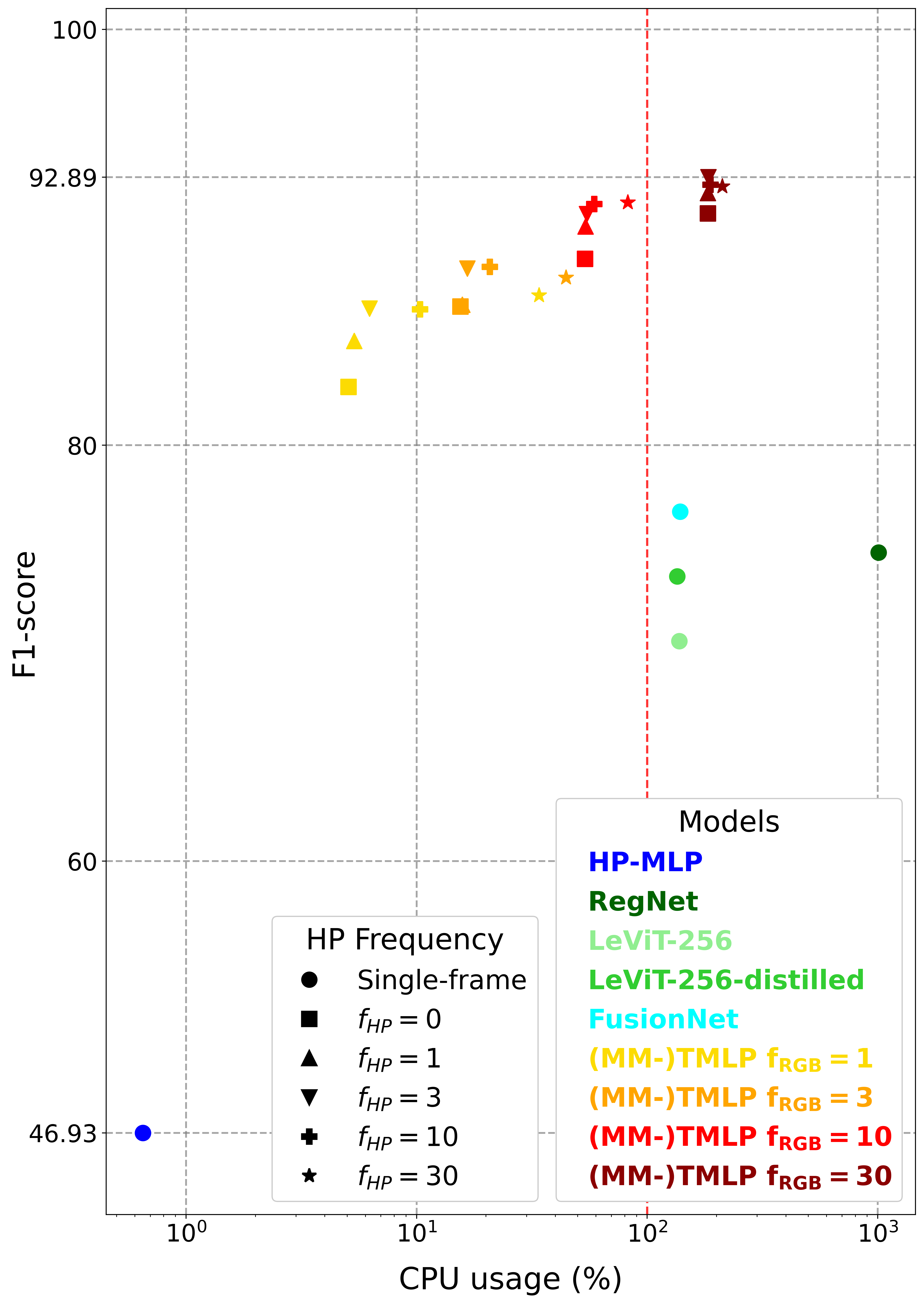}
    \caption{Relationship between F1-score and CPU usage (log scale) for different action prediction models. All the proposed sequence models outperform the single-frame models and highlight how reducing the RGB frequency significantly lowers the CPU usage with limited impact on performance, enabling adaptable design choices for different scenarios. CPU usage is estimated by running inference over a one-second input window.}
    \label{fig:cpu-usage-vs-f1-score}
\end{figure}

We present a combined analysis of model performance and CPU usage. The results are displayed in Figure \ref{fig:cpu-usage-vs-f1-score}. Reported values refer to the macro F1-score. All sequence models were trained on 2-second sequences, corresponding to 60 time steps.

\subsection{Single-frame models}

We evaluated the performance of single-frame models. While RegNet \cite{radosavovic2020designing} achieved a higher F1-score than LeViT-256, its higher CPU usage makes it less suitable for resource-limited environments.

Distillation \cite{hinton2015distilling} helped narrow the performance gap between LeViT-256 and RegNet-12GF while maintaining the same resource usage as the original LeViT-256. Therefore, we selected the LeViT-256-distilled model as the RGB feature extractor.

Additionally, we experimented with an MLP model referred to as HP-MLP. While effective for verb classification, its lower action prediction F1-score suggests that hand pose data lacks object-related cues. Here, verb prediction refers to identifying the general action type (e.g., ``grab book'' and ``grab cappuccino'' are both classified as ``grab''). Despite its limitations, HP-MLP's low CPU usage makes it ideal for practical deployment.

Lastly, we explored single-frame fusion using the FusionNet model, which combines the outputs of the two feature extractors. FusionNet outperformed LeViT-256-distilled, highlighting the complementary nature of these modalities in the single-frame setting.

\subsection{RGB-only Sequence Models}

We experimented with sequence models processing RGB frames (square markers in Figure \ref{fig:cpu-usage-vs-f1-score}). We observe a steady F1-score decline from the model with $f_{\text{RGB}}=30$ Hz to the model with $f_{\text{RGB}}=1$ Hz. However, CPU usage drops more significantly, roughly by a factor of three with each threefold reduction in frequency. These results indicate that RGB frequency can be adjusted to fit resource constraints with minimal performance loss.

\subsection{Multimodal Sequence Models}

We evaluated MM-TMLP models across various combinations of RGB and hand pose sampling frequencies. A comparison with RGB sequence models shows that, for a fixed $f_{\text{RGB}}$, incorporating the hand pose stream consistently improves performance, regardless of the hand pose sampling frequency $f_{\text{HP}}$.

For a fixed $f_{\text{HP}}$, the F1-score declines more sharply as $f_{\text{RGB}}$ decreases, though the hand pose stream helps mitigate this drop. Additionally, the performance gap between the best (MM-TMLP with $f_{\text{RGB}}=30$ Hz and $f_{\text{HP}}=3$ Hz) and worst (MM-TMLP with $f_{\text{RGB}}=1$ Hz and $f_{\text{HP}}=1$ Hz) models is smaller than in RGB-only settings.

The CPU usage primarily depends on $f_{\text{RGB}}$, decreasing roughly threefold with each reduction in $f_{\text{RGB}}$, which is consistent with RGB-only models. Adding the hand pose modality increases CPU usage, particularly at higher $f_{\text{HP}}$ values, as expected.

These trends highlight a key insight: configurations using 10 Hz RGB and 10–30 Hz hand pose frequencies achieve nearly the same F1-score as the full 30 Hz RGB and 30 Hz hand pose setup, while reducing CPU usage by approximately 3×. This demonstrates that significant efficiency gains can be achieved with minimal performance loss, enabling flexible deployment depending on available computational resources.
\section{Conclusions}

In this work, we systematically explored the tradeoffs between accuracy and CPU usage in egocentric action prediction for resource-constrained settings. Our results provide a complete characterization of these trade-offs. In particular, they highlight that lowering the sampling rate for modalities with heavier CPU usage such as RGB (e.g., decreasing $f_{\text{RGB}}$ from 30 Hz to 10 Hz while keeping $f_{\text{HP}}=30$ Hz) leads to only minor accuracy degradation while significantly improving CPU efficiency.

Future work includes enhancing both the efficiency and accuracy of EAR models. One promising direction is to optimize the RGB stream using lightweight architectures or quantization to further reduce computational cost. Another key objective is to develop a large, specialized dataset for egocentric single-frame action recognition, addressing the current data gap in this domain. Additionally, exploring new modalities---such as audio, gaze tracking, and head pose---may provide further opportunities for performance improvement.
{
    \small
    \bibliographystyle{ieeenat_fullname}
    \bibliography{main}
}

\end{document}